\pdfoutput=1

\documentclass[11pt]{article}

\usepackage{EACL2023}

\usepackage{times}
\usepackage{latexsym}

\usepackage[T1]{fontenc}

\usepackage[utf8]{inputenc}

\usepackage{microtype}

\usepackage{inconsolata}

\usepackage{times}
\usepackage{latexsym}

\usepackage[T1]{fontenc}

\usepackage[utf8]{inputenc}

\usepackage{microtype}
\usepackage{times}
\usepackage{epsfig}
\usepackage{graphicx}
\usepackage{subfigure}
\usepackage{amsmath}
\usepackage{amssymb}
\usepackage{multirow}
\usepackage{booktabs}       
\usepackage{soul}

\usepackage{latexsym,bm}
\title{Multimodal Graph Transformer for Multimodal Question Answering}

\author{Xuehai He \\
  UC Santa Cruz \\
  \texttt{xhe89@ucsc.edu} \\\And
  Xin Eric Wang \\
  UC Santa Cruz \\
  \texttt{xwang366@ucsc.edu} \\}

\begin{document}
\maketitle
\begin{abstract}
Despite the success of Transformer models in vision and language tasks, they often learn knowledge from enormous data implicitly and cannot utilize structured input data directly. On the other hand, structured learning approaches such as graph neural networks (GNNs) that integrate prior information can barely compete with Transformer models.
In this work, we aim to benefit from both worlds and propose a novel Multimodal Graph Transformer for question answering tasks that requires performing reasoning across multiple modalities. We introduce a graph-involved plug-and-play quasi-attention mechanism to incorporate multimodal graph information, acquired from text and visual data, to the vanilla self-attention as effective prior. 
In particular, we construct the text graph, dense region graph, and semantic graph to generate adjacency matrices, and then compose them with input vision and language features to perform downstream reasoning. Such a way of regularizing self-attention with graph information significantly improves the inferring ability and helps align features from different modalities.
We validate the effectiveness of Multimodal Graph Transformer over its Transformer baselines on GQA, VQAv2, and MultiModalQA datasets.
\end{abstract}

\section{Introduction}
\begin{figure}[t]
	\begin{center}
 	\includegraphics[width = \columnwidth]{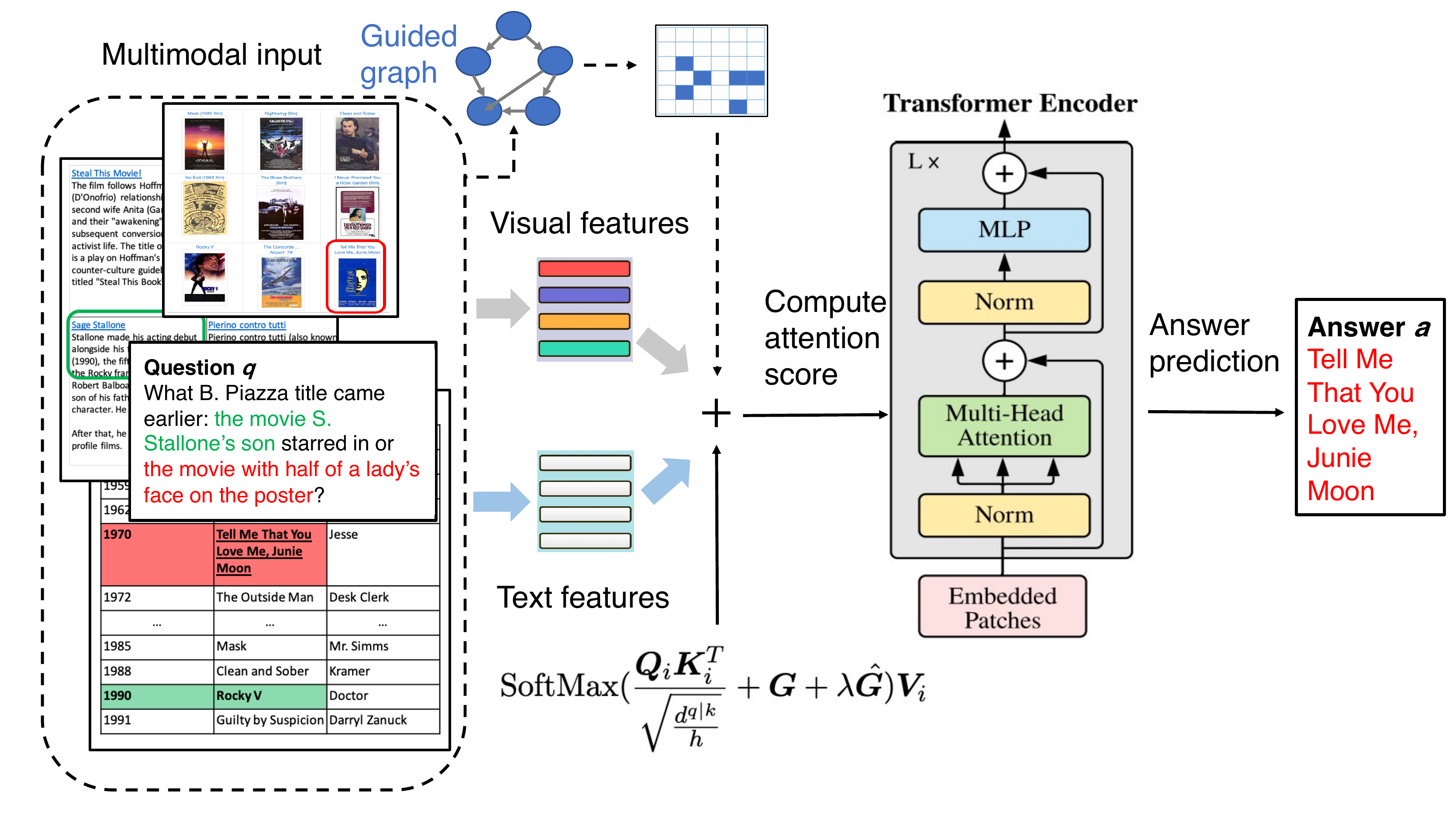}
 	\caption{Overview of Multimodal Graph Transformer. It takes visual features, text features, and their corresponding generated graphs as inputs. The generated graph is first converted to an adjacency matrix to induce the mask matrix $\bm{G}$.  The modified quasi-attention score in the Transformer is computed to infer the answer. In the formular, $\bm{G}$ is the graph-induced matrix constructed by concatenating adjacency matrices both from the vision and the language end. $\hat{\bm{G}}$ is the trainable bias. The input features from different modalities are fused along with graph info to perform downstream reasoning. 
	}\label{fig:teaser}
	\end{center}
 \end{figure}
 
A myriad of complex real-world tasks require both prior knowledge and reasoning
intelligence~\cite{reason1,reason2}. These days, vision-and-language reasoning tasks such as as vision question answering (VQA)~\cite{vqa} and multimodal question answering (MultiModalQA)~\cite{multimodalqa} post further needs for integrating structured info from different input modalities and thus perform reasoning. Towards this, two questions yield: What is the best way to integrate prior knowledge and reasoning components from multiple modalities in a single model? How would such an integration lead to
accurate models, while being more computationally efficient and allowing for significantly more interpretability? Such questions are important
to address when scaling reasoning systems to real-world use cases.

These years, there are a spectrum of methods in the literature exploring different ways of integrating structured prior information. Graph neural networks (GNNs)~\cite{graph_neural_networks}, 
have been widely used in representation learning on graphs.
Some experts tried to investigate the embedding of the structured information by resorting to them. However, GNNs are inefficient~\cite{graph_neural_networks} and they can barely compete with Transformer models. Besides, most GNNs are designed to learn node
representations on fixed and homogeneous graphs. 
Thereby, it is suboptimal to operate GNNs on vision-and-language tasks such as visual question answering (VQA), where graphs encountered in these problems (e.g. scene graphs) can be more complex;
Alternatively, knowledge graphs (KGs), such as Freebase~\cite{knowledge_graph}, represent world-level factoid information of entities and their relations in
a graph-based format, surfaced these years. They have been successfully used
in vision and language applications including VQA~\cite{okvqa}.  However, they have not been dedicated to be applied to our scenario, more concretely, we aim at filling the gap of capturing prior knowledge in Transformer models.

To mitigate deficiencies of the existing methods, this paper proposes a novel plug-and-play graph-involved Transformer-based method for multimodal question answering tasks. Our method is {\em Multimodal Graph Transformer} in the sense that it is built upon the well-established Transformer~\cite{transformers} backbone, albeit with several key fundamental differences. 
First, we introduce a systematic scheme to convert text graphs, dense region graphs, and semantic graphs from vision and language tasks to adjacency matrices to use in our method. 
Second, instead of directly computing the attention score, we learn the newly proposed quasi-attention score with graph-induced adjacency matrices live at its heart, to signify the importance of learning relative importance as a highly effective inductive bias for computing the quasi-attention score.  
Third, different from previous Transformer methods, where self-attention
are fully learned from data, we switch gears to introduce the graph-structured information in the self-attention computation to guide the training of Transformers as shown in Figure~\ref{fig:teaser}.

The main contributions are summarized below:
\begin{itemize}
\item We propose a novel Multimodal Graph Transformer learning framework that combines multimodal graph learning from unstructured data with Transformer models.
\item We introduce a modular plug-and-play graph-involved quasi-attention mechanism with a trainable bias term to guide the information flow during training.
\item The effectiveness of the proposed methods is empirically validated on GQA, VQA-v2, and MultiModalQA tasks. 
\end{itemize}

\section{Related Works}
\subsection{Multimodal question answering}
Visual Question Answering (VQA)\cite{vqa} has been a prominent topic in the field of multimodal question answering, garnering significant attention and advancing significantly since the introduction of the first large-scale VQA dataset by\citet{vqa}. To answer VQA questions, models typically leverage variants of attention to obtain a representation of the image that is relevant to the question~\cite{andreas2016neural, Yang2015StackedAN, xu2016ask, mcb, lu2016hierarchical}. A plethora of works~\cite{liang2021graghvqa, hudson2018compositional, yi2018neural, xiong2016dynamic, BAN, teney2017graph} have attempted to enhance the reasoning capability of VQA models, with~\citet{teney2017graph} proposing to improve VQA using structured representations of the scene contents and questions. They developed a deep neural network that leverages the structure in these representations and builds graphs over scene objects and question words. The recent release of MultiModalQA~\cite{multimodalqa}, a dataset that demands joint reasoning over texts, tables, and images, has received widespread attention. However, similar to VQA, existing MultiModalQA methods have not fully utilized structured information from the input concepts. To address this, we propose a combination of multimodal graph learning and Transformer models to improve question answering across inputs from multiple different modalities.

\subsection{Attention mechanisms}
The attention mechanism~\cite{show_attend_tell, attention, bert}, has dramatically advanced the field of representation learning in machine learning. The attention mechanism is introduced in~\citet{vaswani2017attention} and widely used in language tasks (i.e., abstract summarization~\cite{xu2020self}), machine translation~\cite{bahdanau2014neural}, reading comprehension~\cite{dai2020funnel}, question answering~\cite{min2019knowledge}, etc. ~\citet{sgnet} proposes using syntax to guide the text modeling by incorporating explicit syntactic constraints into attention mechanisms. Meanwhile, it has seen increasing application in multimodal tasks~\cite{li2020oscar, nam2017dual, lu2016hierarchical}, where it is usually used for learning of interactions between multiple inputs.  Following their success, Transformer models have also shown impressive
results on several vision-and-language tasks~\cite{chen2019uniter, hu2020iterative, he2022parameter, videobert}. ~\citet{graph_transformer} proposes Graph Transformer Networks (GTNs) that can generate new graph structures and learn effective node representation on the new graphs in an end-to-end fashion.  Different from these works, our work incorporates graph information from different modalities into the Transformer to improve the reasoning ability.

\subsection{Exploiting graphs in multimodal reasoning}
Considering that graph priors can
transfer commonalities and mitigate the gap between visual and language domains, researchers explore how to use
graphs~\cite{graph_vl1, graph_vl2} properly in both tasks. In recent years, many classes of GNNs have been developed for both tasks which are divided into two approaches: spectral~\cite{bruna2013spectral} and non-spectral methods~\cite{chen2018fastgcn}. Graphs can also be transferred into latent variables by GCN~\cite{gcn1, gcn2},
which can be directly utilized by models. However, the need for
aligning graph priors from different modalities to do reasoning limits the use of graph priors. Our work addresses this problem via the graph-involved quasi-attention mechanism.

\subsection{Pretraining}
Pretrained models in computer vision~\cite{VGG, resnet} and NLP~\cite{bert, yang2019xlnet, liu2019roberta},
have achieved state-of-the-art performances in many downstream tasks~\cite{thongtan-phienthrakul-2019-sentiment, white2017inference, comclip, image_text_matching1, image_text_matching2}.
Other pretrained models~\cite{lu2019vilbert, videobert} based on BERT~\cite{bert} and ViLT~\cite{vilt} also demonstrate their effectiveness on downstream vision-language tasks. Recent works on vision-language pretraining such as OSCAR~\cite{li2020oscar} perform cross-modal alignment in their visual-language pretraining models. Likewise, our proposed method includes cross-modality alignment, which is critical for reasoning.
Our proposed modular plug-and-play graph-involved quasi-attention mechanism is also model-agnostic and can be also applied to other pretrained Transformer-based vision and language models.

\begin{figure*}[htbp]
	\begin{center}
 	\includegraphics[width = 2\columnwidth]{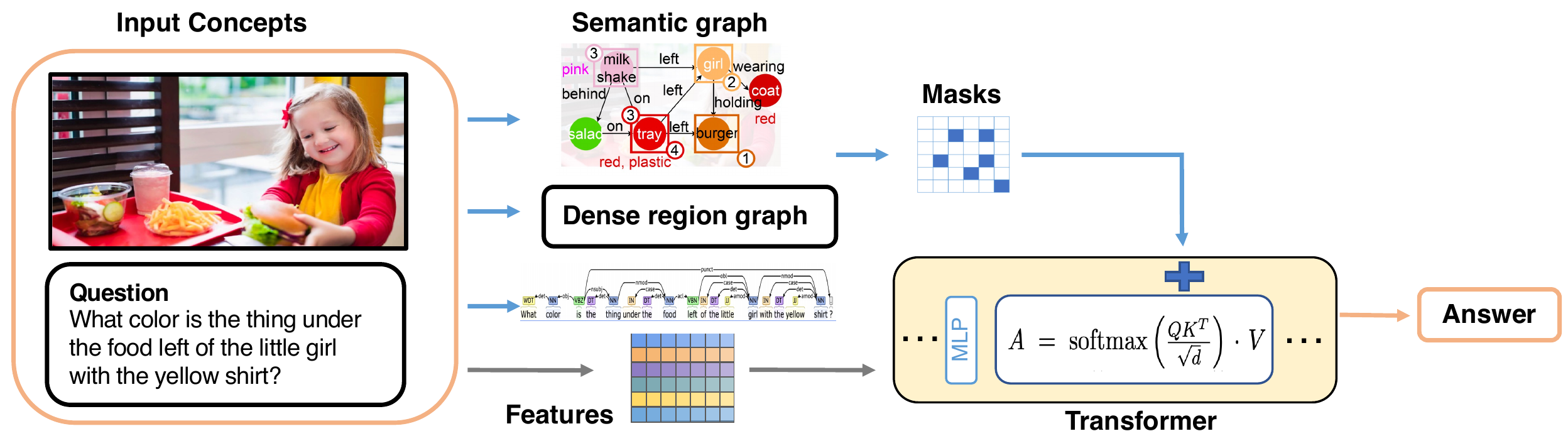}
 	\caption{The figure illustrates the overall framework of our Multimodal Graph Transformer. The input from different modalities are processed and transformed into corresponding graphs, which are then converted into masks and combined with their features to be fed into Transformers for downstream reasoning. In detail, semantic graphs are created through scene graph generation methods, dense region graphs are extracted as densely connected graphs, and text graphs are generated through parsing. 
	}\label{fig:overview}
	\end{center}
 \end{figure*}

\section{Multimodal Graph Transformer}

\subsection{Background on Transformers}
The Transformer layer~\cite{vaswani2017attention} consists of two modules: a multi-head attention and a feed-forward network (FFN). Specifically, each head is represented by four main matrices: the query matrix $\boldsymbol{W}_{i}^{q} \in$ $\mathbb{R}^{d^{m} \times d^{q} / h}$, the key matrix $\boldsymbol{W}_{i}^{k} \in \mathbb{R}^{d^{m} \times \frac{d^{k}}{h}}$, the value matrix $\boldsymbol{W}_{i}^{v} \in \mathbb{R}^{d^{m} \times \frac{d^{v}}{h}}$, and the output matrix $\boldsymbol{W}_{i}^{o} \in \mathbb{R}^{\frac{d^{v}}{h} \times d^{o}}$, and takes the hidden states $\boldsymbol{H} \in \mathbb{R}^{l \times d^{m}}$ of the previous layer as input, where $d$ denotes the dimension of the model, $h$ represents the number of head, and $i$ denotes the index of layer number. The output of attention is given by:\\
\begin{equation}
\boldsymbol{Q}_{i}, \boldsymbol{K}_{i}, \boldsymbol{V}_{i}=\boldsymbol{H} \boldsymbol{W}_{i}^{q}, \boldsymbol{H} \boldsymbol{W}_{i}^{k}, \boldsymbol{H} \boldsymbol{W}_{i}^{v}
\end{equation}
\begin{equation}
\operatorname{Attention}\left(\boldsymbol{Q}_{i}, \boldsymbol{K}_{i}, \boldsymbol{V}_{i}\right)=\operatorname{SoftMax}\left(\frac{\boldsymbol{Q}_{i} \boldsymbol{K}_{i}^{T}}{\sqrt{\frac{d^{q, k}}{h}}}\right) \boldsymbol{V}_{i} \; 
\end{equation}
\begin{equation}\boldsymbol{H}_{i}=\operatorname{Attention}\left(\boldsymbol{Q}_{i}, \boldsymbol{K}_{i}, \boldsymbol{V}_{i}\right) \boldsymbol{W}_{i}^{o}
\end{equation}
where $\boldsymbol{Q}_{i} \in \mathbb{R}^{l \times \frac{d^{q}}{h}}, \boldsymbol{K}_{i} \in \mathbb{R}^{l \times \frac{d^{k}}{h}}, \boldsymbol{V}_{i} \in$ $\mathbb{R}^{l \times \frac{d^{v}}{h}}$ are obtained by the linear transformations of $\boldsymbol{W}_{i}^{q}, \boldsymbol{W}_{i}^{k}, \boldsymbol{W}_{i}^{v}$ respectively. $\operatorname{Attention}(\cdot)$ is the scaled dot-product attention operation. Then output of each head is transformed to $\boldsymbol{H}_{i} \in \mathbb{R}^{l \times d^{o}}$ by $\boldsymbol{W}_{i}^{o} .$

\subsection{Framework overview} 
The entire framework of the proposed Multimodal Graph Transformer method is depicted in Figure~\ref{fig:overview}.  Without loss of generality, we assume the end task is VQA in the following discussion while noting that our framework can be applied to other vision-language tasks, such as multimodal question answering.

Given the input images and questions, the framework first constructs three graphs, including the semantic graph, dense region graph, and text graph, which will be described in more detail in the following sections. The graph $G=(\mathcal{V}, \mathcal{E})$, where $\mathcal{V}$ represents the set of nodes in the graph and $\mathcal{E}$ represents the edges connecting them, is fed into Transformers to guide the training process.

\subsection{Multimodal graph construction}
We build three types of graphs and feed them into Transformers: \emph{text graph}, \emph{semantic graph}, and \emph{dense region graph}. We now introduce them in detail.
\paragraph{Text graph} 
\begin{figure}[htbp]
	\begin{center}
 	\includegraphics[width = \columnwidth]{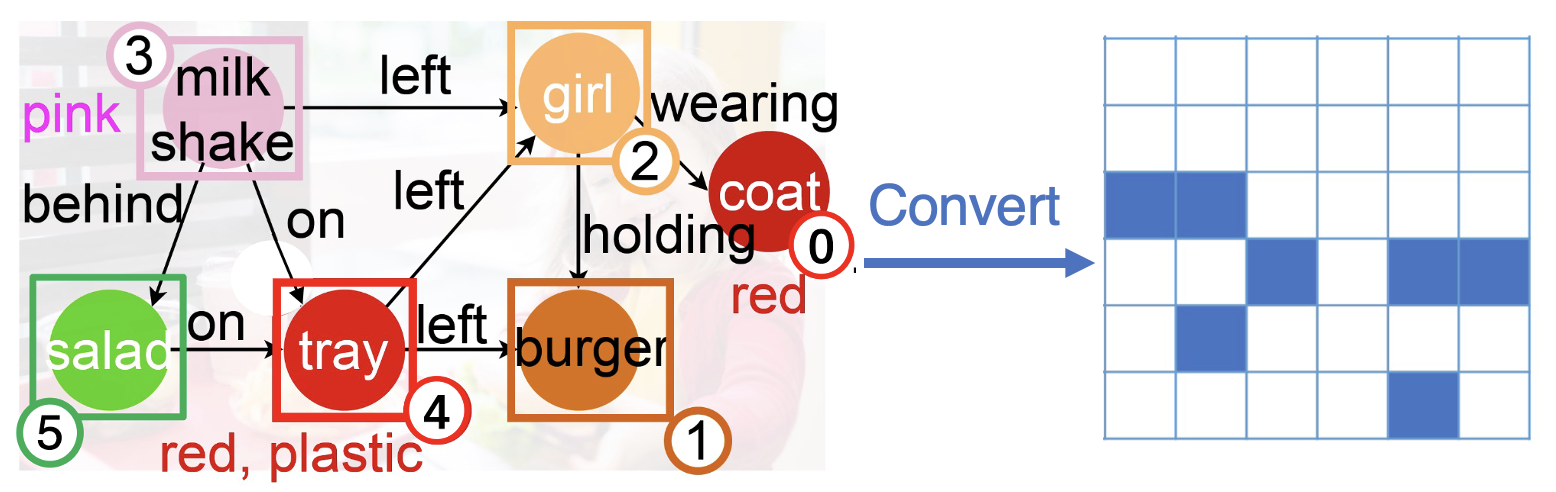}
 	\caption{ The naive demonstration of converting a semantic graph into an adjacency matrix. Cells in blue means `0's for that element in the graph matrix, while white ones means `-inf's. We employ the matrix as the mask when computing the quasi-attention. 
	}\label{fig:convert}
	\end{center}
 \end{figure}
The task of Visual Question Answering involves a combination of an image, a question, and its corresponding answer. To process the question, we extract the entities and create a text graph representation. We then build the graph $G=(\mathcal{V}, \mathcal{E})$ as shown in the left of Figure~\ref{fig:overview}. The set of nodes, $\mathcal{V}$, represents the entities and the set of edges, $\mathcal{E}$, represents the relationships between the pairs of entities. This results in:

\begin{itemize}
\item A set of $N$ entities, each represented by a vector of token embeddings, that constitute the nodes of the graph.
\item A set of pairwise relations between entities, forming the edges of the text graph. The relationship between entities $i$ and $j$ is represented by a vector $e_{i j}$ which encodes the relative relationships.
\end{itemize}

\paragraph{Semantic graph}
In tasks such as multimodal question answering, there might be additional inputs in the form of tables or lengthy paragraph sentences. To handle these inputs, a linear representation of the table can be created and a semantic graph can be constructed using a similar approach. They are processed using the scene graph parser~\cite{zhong2021learning}, which transforms the text sentence into a graph of entities and relations, as depicted in Figure~\ref{fig:convert}. The output of the scene graph parser includes:

\begin{itemize}
\item A set of $N$ words that constitute the nodes of the semantic graph, where $N$ is the number of parsed words in the texts.
\item A set of possible pairwise relations between words, such as "left" and "on" as shown in Figure~\ref{fig:convert}, which constitute the edges of our graph. An edge between words connecting $j$ to $i$ is represented by $e_{i j}$, namely, the connectivity is indicated as: $e_{i j}= \begin{cases}0, & i, j \text{ \ not connected} \\ 1, & i, j \text{ \ connected} \end{cases}$. 
\end{itemize}

\paragraph{Dense region graph}
The visual features are extracted by slicing the input images into patches and flattening them. A dense region graph $G=(\mathcal{V}, \mathcal{E})$ is then converted into masks, with $\mathcal{V}$ being the set of extracted visual features and $\mathcal{E}$ being the set of edges connecting each feature node, following the method described in~\cite{vilt}. This results in a graph that is nearly fully connected.

The resulting three graphs are then transformed into adjacency matrices, where the elements are either -$\infty$ or zero. The conversion process is depicted in Figure~\ref{fig:convert} using the semantic graph as an example. These adjacency matrices are used inside the scaled dot-product attention to control the flow of information, by masking out (setting to $-\infty$) the values.

\subsection{Graph-involved quasi-attention}
\begin{figure}[tbp]
	\begin{center}
 	\includegraphics[width = \columnwidth]{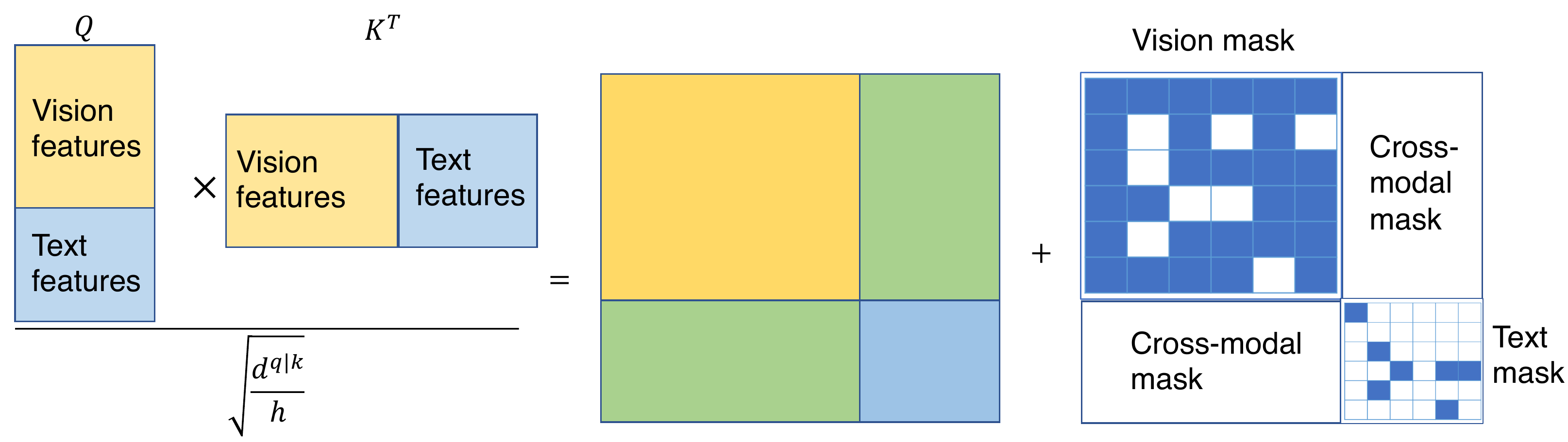}
 	\caption{ A naive demonstration of adding the graph-induced mask while computing the quasi-attention when the inputs are from two modalities. The visual mask is the mask converted from the dense region graph and the text mask is converted from the text graph. The cross-modal mask, which is always set as an all-zero matrix, is imposed to encourage the model to learn the cross-attention between the image features and text features, thus facilitating the alignment across them.
	}\label{fig:graph_mask}
	\end{center}
 \end{figure}

In order to effectively utilize structured graph knowledge in our self-attention computation, we incorporate the graph as an extra constraint in each attention head by converting it into an adjacency matrix. The graph matrix, denoted as $\bm{G}$, is constructed by combining various masks. An illustration of this process can be seen in Figure~\ref{fig:graph_mask}. The visual mask is generated from the dense region graph, while the text mask is derived from the text graph. Additionally, the cross-modal mask is set to an all-zero matrix to encourage the model to learn the cross-attention between visual and text features, thereby promoting alignment across the different modalities.

Within the context of adding graph information,
when vision graph mask and text graph mask are concatenated and aligned with image and text features, we believe that a more flexible masking-out mechanism is beneficial, rather than keeping a single constant mask matrix inside the Softmax operation. Drawing insights from ~\citet{liu2021swin}, where they include a relative position bias to each head in computing similarity, we also intuitively parameterize a trainable bias $\hat{\bm{G}}$ and involve it in the training process. Finally, we compute the quasi-attention as follows:
\begin{equation}
\begin{aligned}
\operatorname{Attention}=\operatorname{SoftMax}(\frac{\boldsymbol{Q}_{i}\boldsymbol{K}_{i}^{T}}{\sqrt{\frac{d^{q \mid k}}{h}}}+\bm{G}+\lambda \hat{\bm{G})  } \boldsymbol{V}_{i} ,
\end{aligned}
\label{eq_additive_term}
\end{equation}

 where $\lambda$ is the tradeoff hyper-parameter that controls the contribution of
$\bm{\hat{G}}$, and $\bm{G}$ is our graph-induced matrix constructed by concatenating a graph matrix both from the vision and the language end.
Here for clear clarification, we use $\bm{G}$ and $\hat{\bm{G}}$ to
distinguish the graph matrices fixed and trainable, respectively. During training, $\bm{G}$ is frozen as before and does not receive gradient updates, while $\hat{\bm{G}}$ contains trainable parameters.  

We now introduce the motivation behind adding two types of graph matrices. We perform the masking process by adding $\bm{G}$ when computing the quasi-attention because 
it can be interpreted as a form of attentional pooling (learning to align), in
which each element of $\bm{G}$ pools all relevant information across all elements of the relative importance matrix computed by $\left(\frac{\boldsymbol{Q}_{i} \boldsymbol{K}_{i}^{T}}{\sqrt{\frac{d^{q \mid k}}{h}}}\right)$. Hence during fine-tuning, the model ignores redundant features and only focuses on useful information. The mask can also force the model to learn the cross attention between features from the images and questions and perform aligning across them. 
And the trainable bias $\hat{\bm{G}}$ captures information gained during the training process. Such information is valuable for fine-tuning, making the Transformer more robust and helping it gain numerical stability.

\subsection{Training}
The interdependence of output features from various modalities calls for a unified optimization approach for the Transformers in both the visual question answering and multimodal question answering tasks. To accomplish this, we implement a kind of end-to-end training, which ensures the optimality of the models. The final outcome of our models is a classification logit, which is generated by the VQA models that select the best answer from the available candidate answers. To evaluate the accuracy of the models, we compute the cross-entropy loss~\cite{cross_entropy} using the output logits produced by the Transformer. This measure helps us determine the difference between the predicted class probabilities and the actual class labels.

\begin{table*}[ht]
\small
 \caption{Accuracy (\%) comparison of different methods on the VQA task. Ours has the second best performance and is comparable to state-of-the-art methods. After applying our proposed quasi-attention mechanism and exploiting the use of graphs, there is also a 2\% improvement of overall accuracy on the LXMERT baseline, suggesting the generalization ability of our method.}
  \centering
  \begin{tabular}{ccccc}
    \toprule
     Dataset &Method & Open questions & Binary questions & Overall accuracy\\
    \midrule
   \multirow{5}*{GQA}& LXMERT~\cite{lxmert} & -&-&60.0  \\   & LXMERT w/ Graph~\cite{lxmert} & -&-&61.4  \\
   & HANs~\cite{kim2020hypergraph} & -&-& 69.4 \\
   & NSM~\cite{hudson2019learning} & 49.3& 78.9& 63.2  \\
   & OSCAR~\cite{li2020oscar} & -&-&61.6 \\
   & VinVL~\cite{zhang2021vinvl} & - &-& 65.1 \\
   & Multimodal Graph Transformer (Ours) &  59.4& 80.5&68.7 \\
    \midrule
\multirow{6}*{VQA v2}& LXMERT~\cite{lxmert} & -&-&72.4  \\
& HANs~\cite{kim2020hypergraph} & -&-&65.1  \\
& NSM~\cite{hudson2019learning} & -&-&63.0  \\
& OSCAR~\cite{li2020oscar} & -&-&73.8 \\
& VinVL~\cite{zhang2021vinvl} & - &-&  76.6 \\
    &Multimodal Graph Transformer (Ours)&66.5&87.0&74.5 \\
    \bottomrule
  \end{tabular}
  \vskip 0.1in
 \label{tab:real_results}
\end{table*}

\section{Experiments}
\subsection{Datasets}

\paragraph{VQA v2}
The VQA v2 dataset~\cite{goyal} extends the VQA~\cite{vqa} dataset to better balance visual and textual information through the collection of complementary images. Each question in VQA v2 is associated with a pair of similar images with different answers, resulting in a total of 1.1 million QA pairs and 204,000 images.
The data split for VQA v2 includes a training set with 83,000 images and 444,000 questions, a validation set with 41,000 images and 214,000 questions, and a test set with 81,000 images and 448,000 questions.
The annotated answers are in natural language, but they are commonly converted to a classification task with 3,129 answer classes. As described by~\citet{bottom}, the model selects the answer to each question from a set of 3,129 most frequent answers. Following this convention, we fine-tune the multimodal graph transformer model on the VQAv2 training and validation sets, while reserving 1,000 validation images and related questions for internal validation.

\paragraph{GQA}
The GQA dataset contains 22M questions over 113K images. The questions in GQA are designed to require multi-hop reasoning to test the reasoning skills of VQA models. GQA greatly increases the complexity of the semantic
structure of questions, leading to a more diverse function
set. The real-world images in GQA also bring in a bigger
challenge in visual understanding. We treat the task as the classification task reffering to the VQA v2 setting.

\paragraph{MultiModalQA}
MultiModalQA (MMQA) contains 29, 918 questions. We split the dataset with reference to the public split. Around 60\% of the questions in MMQA are compositional. The answer for each question can be a single answer or a list of answers.

\subsection{Baselines}
We compare with
four state-of-the-art VQA models:
LXMERT~\cite{lxmert}, NSM~\cite{hudson2019learning}, OSCAR~\cite{li2020oscar}, and VinVL~\cite{zhang2021vinvl}.  

\begin{itemize}
\item LXMERT~\citep{lxmert} designs five pretraining tasks: masked language
modeling, feature regression, label classification,
cross-modal matching, and image question answering to
pretrain a large Transformer model. Towards this, a large-scale Transformer~\citep{vaswani2017attention} 
model is built that consists of three encoders: an object
relationship encoder, a language encoder,
and a cross-modal encoder. 
\item NSM~\cite{hudson2019learning} predicts a probabilistic graph
that represents its underlying semantics and performs sequential reasoning over the graph to traversing its
nodes to make the inference. 
\item OSCAR~\cite{li2020oscar} uses object tags detected in images as anchor points to significantly ease
the learning of alignments, improving previous methods and using self-attention to learn image-text semantic alignments.
\item VinVL~\cite{zhang2021vinvl} developed a new
object detection model to create better visual features of images than previous classical object detection models. 
\end{itemize}

We compare with four baselines introduced in the MultiModalQA paper~\cite{multimodalqa}: Question-only~\cite{questiononly}, Context-only~\cite{questiononly}, AutoRouting, ImplicitDecomp. 
\begin{itemize}
\item Question-only is a sequence-to-sequence
model that directly generates the answer given the question.
\item Context-only first predicts the question type using the classifier and then feed in the relevant context to predict the answer.
\item AutoRouting first determines the modality where the answer is expected to occur, and then runs the corresponding single-modality module. 
\item ImplicitDecomp is a 2-hop implicit decomposition
baseline and so far the state-of-the-art method on the MultiModalQA dataset.
\end{itemize}

\subsection{Implementation details}
The input texts undergo preprocessing using a scene graph parser which extracts entities and their relationships. The text features are obtained through a pre-trained BERT tokenizer, allowing us to extract text spans of individual entities and text spans containing two related entities. As for images, we employ the methods described in~\citet{vit, vilt} to extract visual features and create graph masks. This involves resizing the shorter edge of the input images while preserving the aspect ratio and limiting the longer edge, followed by patch projection and padding for batch training. The resulting patch embeddings are used as inputs along with constructed dense region graph that is densely connected. The Transformer backbone used in this setting is the pretrained VIT-B-32~\cite{vit} version, consisting of 12 layers with a hidden size of $H$ = 768, layer depth of $D$ = 12, patch size of $P$ = 32, a multi-layer perceptron size of 3072, and 12 attention heads. To test this setting, all inputs and graphs are merged and processed by the Transformer backbone, which learns from features from different modalities.

\begin{table}[ht]
\small
 \caption{EM (\%) and F1 (\%) of Multimodal Graph Transformer and its Transformer baseline on questions in MultiModalQA that require reasoning over multiple modalities. We also quote the results from the MultiModalQA~\cite{multimodalqa} paper. Incorporating graph information into the Multimodal Graph Transformer can boost about 2\% F1 and 4\% EM performance.} 
  \centering
  \scalebox{1}{
  \begin{tabular}{ccc}
    \toprule
   Method& EM  & F1
    \\
    \midrule
    Question-only& 16.9&19.5\\
    Context-only&6.6&8.5\\
    \midrule
    AutoRouting& 32.0&38.2\\
    ImplicitDecomp&46.5&51.7\\
     \midrule
     Human&84.8&90.1\\
    \midrule
     \midrule
     Multimodal Transformer w/o Graph & 50.1&56.4\\
     Multimodal Graph Transformer (Ours) &52.1&57.7\\
     
    \bottomrule
  \end{tabular}}
  \vskip 0.1in
 \label{tab:real_results2}
\end{table}

\begin{table*}[ht]
\small
 \caption{Ablation Studies on the GQA and VQA v2 datasets. The figure demonstrates the effectiveness of incorporating graph information into the Transformer architecture through ablation studies performed on the GQA and VQA. The results of these studies clearly indicate that including graph information can lead to an improvement in performance.}
  \centering
  \begin{tabular}{cccccc}
  \toprule
  Dataset  & Method & Open questions & Binary questions & Overall accuracy\\
    \midrule
 \multirow{3}*{GQA} 
   &One-modality Transformer & 47.7 & 78.1&62.7 \\ & Multimodal Transformer w/o Graph &49.9 & 81.0& 65.4 \\
   & Ours & \textbf{60.1} & \textbf{90.2}& \textbf{72.4} \\
   \midrule 
 \multirow{3}*{VQA v2} & One-modality Transformer w/ one Transformer &60.5& 85.4&70.1 \\
   & Multimodal Transformer w/o Graph & 64.8&86.3&72.1  \\
    &Ours  &\textbf{66.7}& \textbf{87.2}&\textbf{74.6} \\
    \bottomrule
  \end{tabular}
  \vskip 0.1in
 \label{tab:real_results1}
\end{table*}
\subsubsection{MultiModalQA}
We further investigate the effectiveness of our proposed method on MultiModalQA~\cite{multimodalqa}, a recently introduced and demanding task that requires joint reasoning across various modalities such as texts, images, tables, etc. We employ a Multimodal Graph Transformer to tackle the task, using the same approach for extracting vision and text features as in VQA. Additional modalities, such as tables, are encoded by linearizing them and utilizing pre-trained models like RoBERTa-large~\cite{liu2019roberta}. After generating text graphs, semantic graphs, and dense region graphs from input questions, text, tables, and images, we feed them along with the extracted features into the Transformer.

\subsection{Results and analysis}
Table~\ref{tab:real_results} presents a comparison of the accuracy of our proposed method on the GQA dataset with previous state-of-the-art methods. Our proposed method ranks second in terms of accuracy and outperforms the third best method by a substantial margin, with an absolute improvement of over 3\% in overall accuracy. The performance of our method is comparable to the state-of-the-art method.

We also conducted experiments on the VQA v2 dataset, and the results are summarized in Table~\ref{tab:real_results} and Table~\ref{tab:real_results1}. As shown, there are significant improvements over methods without graphs, suggesting that incorporating graph information into the Transformer is effective.

Additionally, after incorporating our proposed graph method into LXMERT, we can observe a boost in overall accuracy on the GQA dataset, demonstrating the generalization ability of the proposed method in incorporating graph information into quasi-attention computation.

Table~\ref{tab:real_results2} compares the Exact Match (EM) and average F1 score of our proposed method on the MultiModalQA dataset with the baseline. The results show that our proposed method outperforms the baseline without the aid of graph information, demonstrating the generalization of our method to more complicated vision-and-language reasoning tasks.

\subsection{Ablation studies}
We perform ablation studies to verify the necessity of using two-stream inputs with the help of graphs to deal with input from different modalities, with
GQA dataset as our testing bed. For all experiments, we use the overall accuracy as the evaluation metric.

The results presented in Table~\ref{tab:real_results1} show the superiority of our proposed Multimodal Graph Transformer over the method where a single modality input is fed into a Transformer. Our method, which involves dividing the input streams into two separate parts and processing each part through a Transformer, outperforms the Multimodal Transformer without Graph. This demonstrates the beneficial effect of incorporating graph information into the processing of the input data and performing training. The use of different input features with the help of graphs allows for a better alignment of the information from different modalities, which is reflected in the improved performance of our proposed method.

\subsection{Qualitative results}
\begin{figure}[htbp]
	\begin{center}
 	\includegraphics[width = \columnwidth]{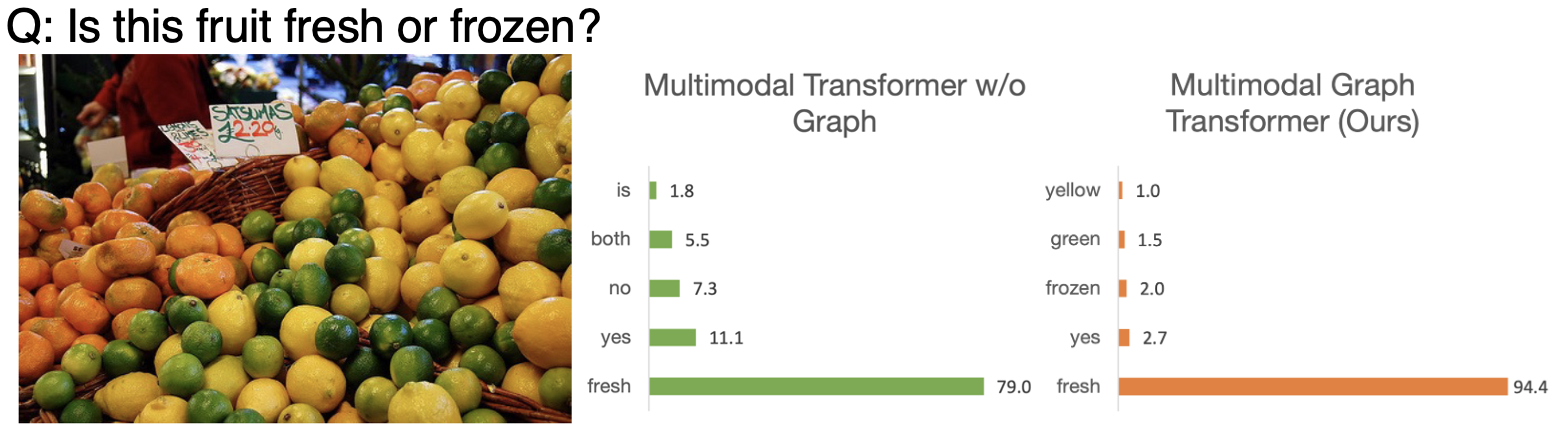}
 	\caption{A qualitative comparison from VQA v2. {\em fresh} is the ground truth. Predictions from the Multimodal Graph Transformer (ours) are more relevant to the contents of the input image and achieve a higher confidence score over the ground truth.}
	\label{fig:Qualitative_Results}
	\end{center}
 \end{figure}

One qualitative example is shown in Figure~\ref{fig:Qualitative_Results}. As can be seen, predictions from Multimodal Graph Transformer are more relevant to contents of the input image as the graph information improves the inferring ability of the Transformer, which further indicates the effectiveness of Multimodal Graph Transformer.

\section{Conclusions}
In this paper, we have presented a novel method to integrate
structured graph information to guide the Transformers training. Our method can model interactions between different modalities and achieves competitive performance on multimodal reasoning tasks such as VQA and MultiModalQA. Experimental results show that our method outperforms many other methods on the GQA dataset. More importantly, the proposed quasi-attention mechanism is model-agnostic and it is possible to apply it to other Transformer-based methods. We will test our methods on other vision-and-language reasoning tasks and include the comparison with existing graph representation learning methods in our future work.

\section{Limitations and Potential Risks}
The Limitations of the proposed Multimodal Graph Transformer include the potential preservation of fairness and bias issues inherent in the pretrained Transformer models, despite the involvement of graph information. Additionally, the integration of graphs may introduce new biases that can further exacerbate the problem. One potential source of bias is the vision-and-language dataset itself, which may favor majority cases and overlook minority cases. Unfortunately, the proposed method is not equipped to address these biases and issues, making further research and consideration crucial when building upon or directly using this method for vision and language tasks.

\bibliography{main}
\bibliographystyle{acl_natbib}

  \begin{table*}[htbp]
 \centering
\caption{Comparison of VQA datasets}
\begin{tabular}{|c|c|c|c|c|c|}
\hline  
&Source of images&\# images& \# QA pairs &Answer type&Evaluation metrics\\
\hline  
DAQUAR&NYU-Depth V2& 1,449&12,468  & Open& Accuracy\&WUPS\\
\hline
VQA&COCO& 204K& 614K & Open/MC & Accuracy\\
\hline
VQA v2 &COCO& 204K& 1.1M & Open/MC & Accuracy\\ 
\hline
COCO-QA & COCO&123K& 118K & Open/MC & Accuracy\\
\hline
CLEVR&Generated& 100K& 999K & Open & Accuracy\\
\hline
GQA&Visual Genome& 113K& 22M & Open & Accuracy\\
\hline
\end{tabular}
\label{tab: dataset}
\end{table*}

\appendix

\section{Appendix}
\label{sec:appendix}

\subsection{Visual Question Answering dataset}
To address the problem of visual question answering, a number of visual question answering datasets have been developed. The comparison of them is shown in Table~\ref{tab: dataset}. The VQA dataset~\cite{vqa} is developed on real images in MS COCO~\cite{coco} and abstract scene images
in~\citet{abstract0,abstract1}. The
question-answer pairs are created by human annotators who are encouraged to ask ``interesting" and ``diverse" questions. VQA v2~\cite{goyal} is extended from the VQA~\cite{vqa} dataset to achieve more balance between visual and textual information by collecting complementary images in a way that each question is associated with a pair of similar images with different answers; In the COCO-QA~\cite{cocoQA} dataset, the question-answer pairs are automatically generated from image captions based on syntactic parsing and linguistic rules; PathVQA~\cite{pathvqad} is a pathology visual question answering dataset proposed to foster the research of medical VQA; DAQUAR~\cite{first} is built on top of the NYU-Depth V2 dataset~\cite{nyu} which contains RGBD images of indoor scenes. DAQUAR consists of (1) synthetic question-answer pairs that are automatically generated based on textual templates and (2) human-created question-answer pairs produced by five annotators;  CLEVR~\cite{clevr} is a dataset developed on rendered images of spatially related objects (including cube, sphere, and cylinder)  with different sizes, materials, and colors. The locations and attributes of objects are annotated for each image. The questions are automatically generated from the annotations; GQA is a new dataset for real-world visual reasoning and compositional question answering, seeking to address key shortcomings of previous VQA datasets. Considering questions in GQA are most objective, unambiguous, compositional, and can be answered by reasoning only on the visual content. We mainly use the GQA dataset in this work as it best fits our goal of reasoning. We also evaluate our methods on the VQA v2 dataset as it is the most common and general VQA dataset so far.
\begin{figure}[htbp]
	\begin{center}
 	\includegraphics[width = \columnwidth]{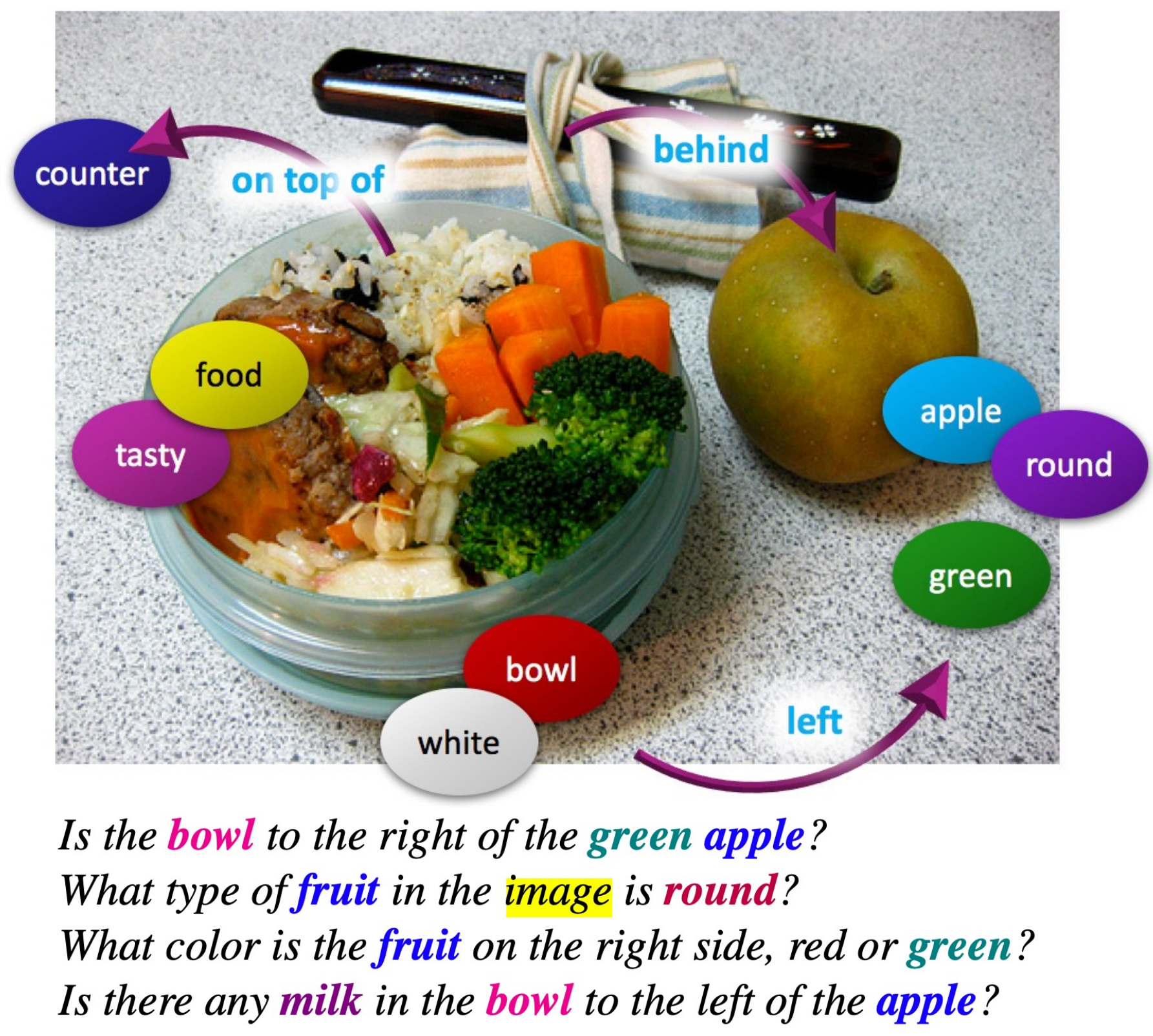}
 	\caption{Examples from the GQA dataset for visual reasoning and compositional question answering.
	}\label{fig:GQA}
	\end{center}
 \end{figure}

\begin{figure}[htbp]
	\begin{center}
 	\includegraphics[width = \columnwidth]{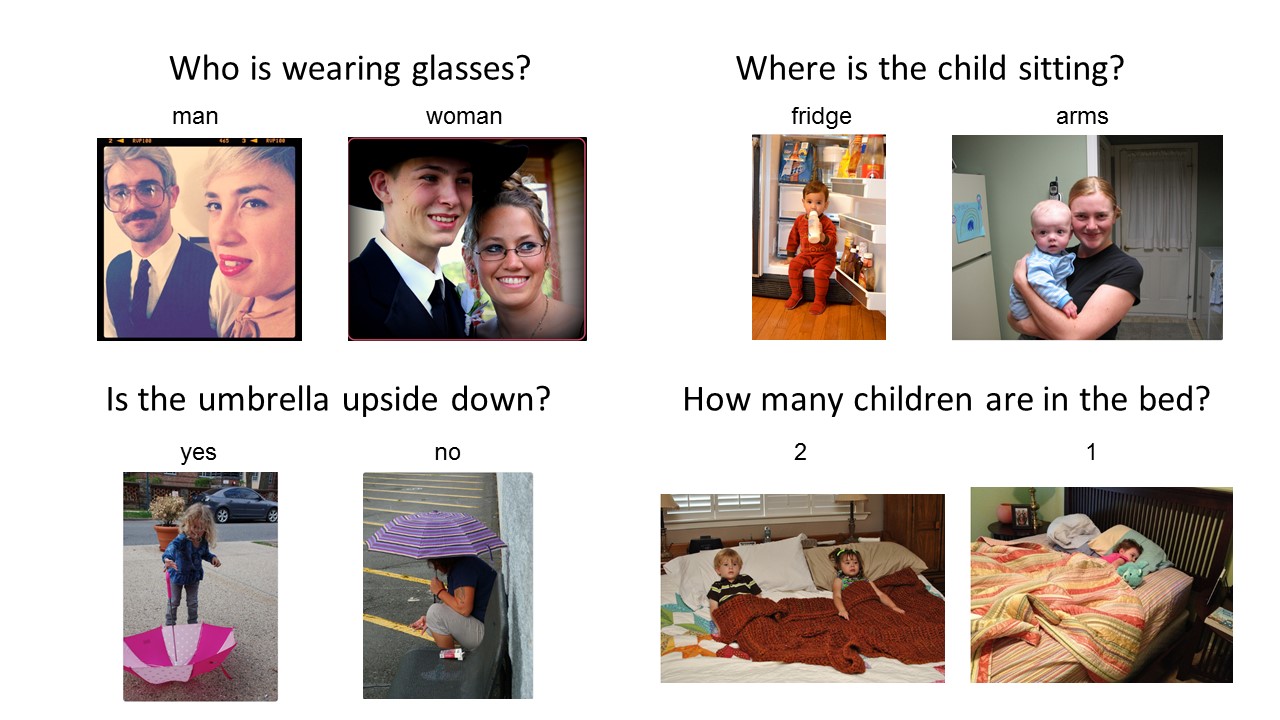}
 	\caption{Examples from the VQA v2 dataset for Visual Question Answering.
	}\label{fig:VQAv2}
	\end{center}
 \end{figure}
 
\end{document}